\documentclass[letterpaper]{article} 
\usepackage[submission]{aaai23}  
\usepackage{times}  
\usepackage{helvet}  
\usepackage{courier}  
\usepackage[hyphens]{url}  
\usepackage{graphicx} 
\usepackage{natbib}  
\usepackage{caption} 
\usepackage{algorithm}
\usepackage{algorithmic}
\usepackage{multirow}
\usepackage{booktabs}
\usepackage{makecell}
\usepackage[switch]{lineno}
\usepackage{newfloat}
\usepackage{listings}
\usepackage{amssymb, bm, amsmath}
\usepackage{subfig,}
\usepackage{xcolor}
\usepackage{eucal}
\usepackage{newfloat}
\usepackage{listings}
\DeclareCaptionStyle{ruled}{labelfont=normalfont,labelsep=colon,strut=off} 
\floatstyle{ruled}
\newfloat{listing}{tb}{lst}{}
\floatname{listing}{Listing}
\title{InterMulti:Multi-view Multimodal Interactions with Text-dominated Hierarchical High-order Fusion for Emotion Analysis}

\author {
    Feng Qiu \textsuperscript{\rm 1},
    Wanzeng Kong \textsuperscript{\rm 2},
    Yu Ding\thanks{* Corresponding author.} \textsuperscript{\rm 1}
}
\affiliations {
    \textsuperscript{\rm 1} Virtual Human Group, Netease Fuxi AI Lab, China\\
    \textsuperscript{\rm 2} College of Computer Science, Hangzhou Dianzi University\\
    qiufeng@corp.netease.com, kongwanzeng@hdu.edu.cn, dingyu01@corp.netease.com
}

\usepackage{bibentry}

\begin{document}

\maketitle

\begin{abstract}
Humans are sophisticated at reading interlocutors' emotions from multimodal signals, such as speech contents, voice tones and facial expressions. However, machines might struggle to understand various emotions due to the difficulty of effectively decoding emotions from the complex interactions between multimodal signals. In this paper, we propose a multimodal emotion analysis framework, InterMulti, to capture complex multimodal interactions from different views and identify emotions from multimodal signals. Our proposed framework decomposes signals of different modalities into three kinds of multimodal interaction representations, including a modality-full interaction representation, a modality-shared interaction representation, and three modality-specific interaction representations. 
Additionally, to balance the contribution of different modalities and learn a more informative latent interaction representation, we developed a novel Text-dominated Hierarchical High-order Fusion(THHF) module. THHF module reasonably integrates the above three kinds of representations into a comprehensive multimodal interaction representation. Extensive experimental results on widely used datasets, \textit{i.e.}, MOSEI, MOSI and IEMOCAP, demonstrate that our method outperforms the state-of-the-art. 
\end{abstract}

\section{Introduction}
Endowing machines with emotional intelligence has been a long-standing goal for engineers and researchers working on artificial intelligence \cite{KDD1,emoAI}. 
In daily conversations, humans are skilled in conveying and understanding emotions through multimodal signals, including spoken words, voice tones and facial expressions. These multimodal signals are supplementary and complementary to each other in expressing emotional information \cite{survy} by skillfully interacting with each other. Their interaction is a sophisticated and delicate process \cite{zadeh2017tensor}. Therefore, it is vital to understand the relationship and interaction among the multimodal signals for emotion analysis.

It has still been challenging to explore complex multimodal interactions for emotion analysis effectively. The exploration aims at learning a fusion representation to offer effective information for emotion analysis. Previous works usually integrate the multimodal information for emotion analysis by direct concatenation \cite{zadeh2017tensor}, attention-assisted fusion \cite{MARN,MFN} or correlation-based fusion \cite{zadeh2017tensor,LMF,high-order}.
Although these methods have attempted to take advantage of complementary information among different modalities, they are often challenged by the gaps among heterogeneous modalities \cite{MISA}. So these methods may be incapable of fully modeling the complex interaction of multiple modalities. The refinement of modeling their complex interactions may contribute to improving emotion analysis.


Later, the recent works capture more complex interactions between different modalities by refining the learning of multimodal fusion representation \cite{MISA, BBFN, Self-MM, MMIM, MTAG}. These works decouple unimodal features into multiple representations via regularizers and then fuse them. They aim to capture independence and correlation between modalities. These works approve the benefit of the decoupling operation in minimizing redundancy between multimodal information. However, the decoupling operation with a regularizer requires several auxiliary orthogonal constraint losses and hyper-parameters balancing these losses, as well as a large quantity of additional trainable parameters. These requirements make the training complex and may limit further improvement. 



To investigate complex multimodal interactions, our framework InterMulti decomposes multiple unimodal features into three kinds of multimodal interaction representations via a nonparametric decoupling method, including: 
\begin{enumerate}
    \item The modality-full representations reserve intact information of the single modality;
    \item The modality-shared interaction representation focuses on the contribution of coordination among multiple modalities;
    \item The modality-specific interaction representations highlight the contribution of specific information from individual modality, excluding common information. 
\end{enumerate}

Importantly, different from previous works, our decoupling method is nonparametric and instance-based. It makes a lower dependency between modality-shared and modality-specific representations at the instance-level than the distribution-based method like MISA \cite{MISA} (Please refer to Table \ref{table:smilarity}). Our instance-based decoupling method requires neither regular loss (e.g. similarity or difference loss) nor a discriminator with additional learnable parameters and hyper-parameters. Therefore, InterMulti is an effective and easy-training framework for modeling complex multimodal interactions.

Then, these three kinds of interaction representations are merged into a comprehensive fusion representation. It has been demonstrated that the text modality plays the most significant role in multimodal emotion analysis \cite{pham2019improving,tsai2019multimodal}. Therefore, we develop a novel module, named Text-dominated Hierarchical High-order Fusion (THHF), to merge the different kinds of interaction representations. THHF guides the model to balance the contribution of different modalities and further unleash the representation capacity of the multimodal interaction for emotion analysis.



To demonstrate the effectiveness of our framework, extensive experiments are conducted on three datasets: MOSI \cite{fukui2016multimodal}, MOSEI \cite{zadeh2018multimodal} and IEMOCAP \cite{busso2008iemocap}. The results show that our model achieves state-of-the-art performance on nearly all benchmarks. Our contributions can be summarised as follows:
\begin{itemize}
\item We propose a novel framework InterMulti that emphasizes and validates the importance of modeling multi-view multimodal interactions for emotion analysis.
\item We develop a nonparametric instance-based decoupling operation and a novel module called Text-dominated Hierarchical High-order Fusion (THHF) to decompose and fuse multimodal features more reasonably and guide the model to explore more informative and effective multimodal interaction representations.


\item Our method achieves state-of-the-art performance on three prevailing benchmarks for emotion analysis.
\end{itemize}

\section{Related Work}
\label{sec:relatedworks}
The goal of learning multimodal fusion representations from different unimodal features is to model the interaction of multimodal features to offer effective information for emotion analysis. As mentioned, most previous works carry out the fusion by directly integrating the three-modal features, highlighting the co-working of the intact unimodal features. This method is named integrating learning in our work. Another work \cite{MISA} first decouples the unimodal features into several intermediate representations and then fuses them. Its underlying idea is that emotion is encoded by not only the coordination of multimodalities but also the separately working. This method is named decoupling learning in our work. The reported integrating learning and decoupling learning methods will be briefly reviewed.


\subsubsection{Integrating learning.}
A straightforward method is the concatenation of unimodal features \cite{MVLSTM, BCLSTM}. Recently, a large number of works have introduced the attention-based mechanism to integrate three-modal features \cite{MARN, MFN, BBFN, MARN, MFN, BBFN, wordsCShift}. 
However, \citet{zadeh2017tensor} indicates that both concatenation and attention-based methods over-fit to a specific modality while ignoring the others and introducing modality bias. Some works attempt to capture more relationships between modalities with various regularizations on the learning of joint representation, such as CCA loss \cite{ICCN}, adversarial mechanism \cite{M2MAdversarial, Adversarial2}, mutual information \cite{MMIM}
and self-supervised manner \cite{Foundintranslation, dynamicfusion, Self-MM}. These methods focus on extracting the common information but ignoring the individual role of each modality in emotion analysis. Other works use tensor fusion to highlight both individual modality representations and their high-order relationships \cite{zadeh2017tensor, high-order, LMF}. 

\subsubsection{Decoupling learning.}
In multimodal emotion analysis, MISA \cite{MISA} is the first work to utilize several orthogonal constraint losses to align features from different modalities before fusion which bridges modality gaps. MISA decouples three unimodal features into a modality-invariant representation and a modality-specific representation, respectively. Their decoupling operation relies on auxiliary orthogonality constraints, as well as additional hyper-parameters to balance these losses. In training, the supervision under several losses makes the training difficult, as the training has to satisfy all the loss functions. In fact, the complex training cannot guarantee the intention of a low similarity between the decoupling representations, which probably restrains the performance of emotion analysis. It is critical to make ease of training but effective performance when decoupling representations.

\let\oldequation\equation
\let\oldendequation\endequation
\renewenvironment{equation}{\linenomathNonumbers\oldequation}{\oldendequation\endlinenomath}
\section{Method}

\begin{figure*}[ht]
\centering
\includegraphics[width=0.98\textwidth]{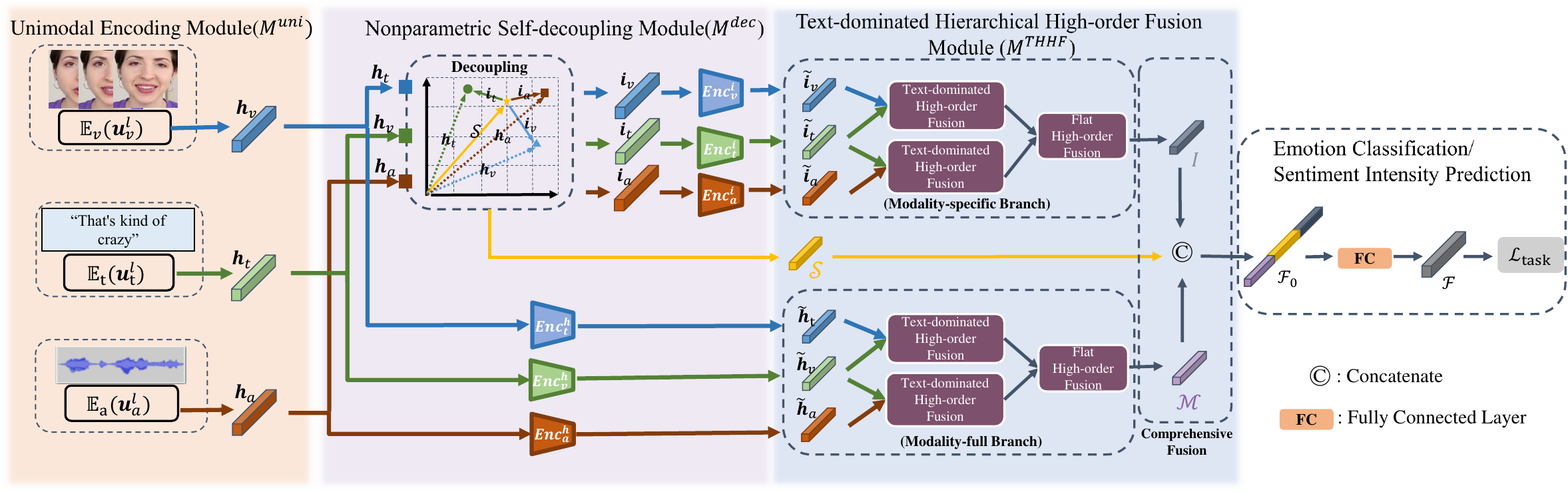}
\caption{Illustration of InterMulti framework. $M^{enc}$ takes the utterance-level unimodal features as input and produces unimodal representations (${\textbf{\textit{h}}_t}$, ${\textbf{\textit{h}}_v}$, and ${\textbf{\textit{h}}_a}$). Then $M^{dec}$ decomposes ${\textbf{\textit{h}}_t}$, ${\textbf{\textit{h}}_v}$ and ${\textbf{\textit{h}}_a}$ and obtain three kinds of interaction representations. Later, the $M^{THHF}$ performs the semantic-based hierarchical fusion from these interaction representations.}
\label{fig:pipeline}
\end{figure*}

This section describes our InterMulti for multimodal emotion analysis from three kinds of temporal input signals of text, audio and vision in a talking-face clip. As illustrated in Figure \ref{fig:pipeline}, our framework InterMulti consists of three modules including {Unimodal Encoding Module} ({$M^{enc}$}), {Nonparametric Decoupling Module} ({$M^{dec}$}) and {Text-dominated Hierarchical High-order Fusion (THHF) Module} ({$M^{THHF}$}).
Given a talking-face clip, we use \textit{$M^{enc}$} to get three unimodal utterances-based features according to their temporal input signals. Then \textit{$M^{dec}$} decomposes these three unimodal features into one modality-shared representation and three modality-specific representations. Additionally, we also get three modality-full representations from utterances-based features. After that, \textit{$M^{THHF}$} is dedicated to modeling the complex interaction across multiple modalities from different views by fusing diverse interaction representations. Lastly, two linear projection layers are introduced to enhance the representation capacity of multimodal interaction representation further. More details will be described as follows.

\subsection{Unimodal Encoding Module}
\label{sec:uniModule}

Given a talking-face clip consisting of three modalities (e.g. text, visual, acoustic), $M^{enc}$ aims to extract unimodal features for each modality. We first obtain the utterance features $\textbf{\textit{u}}_m^l$ (${{m} \in \{{t, v, a} \}}$,  where $t$ refers to text, $a$ to acoustic and $v$ to vision; ${l \in \{1, 2, ..., L\}}$ refers to the index of temporal sequence of modality $m$ where L refers to the total length of the sequence.) by pre-trained unimodal feature extractors (details in Section \ref{sec:experiment}). Then unimodal encoders $\mathbb{E}_m$ compute the utterance-based 64-dimensional unimodal representations $\textbf{\textit{h}}_m$ (i.e. $\textbf{\textit{h}}_t$, $\textbf{\textit{h}}_a$ and $\textbf{\textit{h}}_v$). For each unimodal encoder, we apply two single-layer bidirectional gated recurrent units (BiGRU) \cite{chung2014empirical} followed by a fully connected layer to capture the temporal dependence and cast all the latent vectors into the same length for the convenience of further fusion. The end-to-end learned $\textbf{\textit{h}}_m$ is capable of reducing the modality gaps between heterogeneous signals and aligning them to the same representation space, facilitating their interactions modeling later.

\subsection{Nonparametric Self-decoupling Module}

Taking unimodal representations $\textbf{\textit{h}}_m$ as inputs, \textit{$M^{dec}$} aims to decompose them into different kinds of intermediate representations. The goal of decoupling is to facilitate complex interactions modeling.
To this end, we develop an nonparametric instance-based self-decoupling method to decompose $\textbf{\textit{h}}_m$ into two kinds of representations, including a modality-shared representation $\mathcal{S}$ and three modality-specific representations $\textbf{\textit{i}}_t$, $\textbf{\textit{i}}_a$ and $\textbf{\textit{i}}_v$. $\mathcal{S}$ represents common information across three modalities and $\textbf{\textit{i}}_m$ refers to the specific information for each modality. To facilitate training, our instance-based decoupling requires neither additional trainable parameters nor auxiliary loss functions. 

To be specific, $\mathcal{S}$ is viewed as the average of $\textbf{\textit{h}}_t$, $\textbf{\textit{h}}_a$ and $\textbf{\textit{h}}_v$. Then $\textbf{\textit{i}}_t$, $\textbf{\textit{i}}_a$ and $\textbf{\textit{i}}_v$ are computed by subtracting $\mathcal{S}$ from  $\textbf{\textit{h}}_t$, $\textbf{\textit{h}}_a$ and $\textbf{\textit{h}}_v$, respectively. The subtraction moves out the shared information from $\textbf{\textit{h}}_m$ and reserves the specific information for each modality. This nonparametric method obtains three modality-specific representations with low dependence between different modalities. Otherwise, we would obtain distribution-based low dependence between different modalities, which cannot guarantee low dependence for samples far from the center of the data distribution (Please refer to Table \ref{table:smilarity}). Our nonparametric self-decoupling operation is formulated as follows:
\begin{equation}
\begin{array}{cc}
\mathcal{S} &= \frac{1}{3} \sum\limits_{m} {\textbf{\textit{h}}_m} ; \quad  {m \in \{{t, v, a} \}}  \\ 
{\textbf{\textit{i}}_m} &= \textbf{\textit{h}}_m - \mathcal{S}; \quad  {m \in \{{t, v, a} \}} \\
\end{array}
\end{equation}

To make modality-specific representations more compact, a fully connected layer is introduced to transform each modality-specific representation $\textbf{\textit{i}}_m$ into the 16-dimensional representation $\tilde{\textbf{\textit{i}}}_m$, respectively. The end-to-end training allows $\tilde{\textbf{\textit{i}}}_m$ to learn appropriate modality-specific representation spaces with the guide of the nonparametric decoupling operation and the trainable fully connected layer. It prompts instance-based cross-modality independence and modality-specific representation capacity.

Additionally, besides the modality-shared ($\mathcal{S}$) and modality-specific ($\tilde{\textbf{\textit{i}}}_m$) representations, we still retain the modality-full features $\textbf{\textit{h}}_m$ with all the information of unimodal signals as information supplement. In order to reduce the number of parameters and prepare for further fusion, we first reduce the dimension of $\textbf{\textit{h}}_m$ from 64 to 16 through a fully-connected layer and get the representations $\tilde{\textbf{\textit{h}}}_m$.

\subsection{Text-dominated Hierarchical High-order Fusion Module}
According to the semantic distinction, the above representations are classified into three groups: the full unimodal representations (${\tilde{\textbf{\textit{h}}}}_m$), the modality-specific representations ($\tilde{\textbf{\textit{i}}}_m$), and the modality-shared representation ($\mathcal{S}$).
\textit{$M^{THHF}$} aims at reasonably merging diverse types of representations from different views for emotion analysis by a hierarchical structure. Considering the modality imbalance during multimodal fusion, we further introduce the Text-dominated mechanism to guide the network to explore more effective and informative interaction representations and capture their modality-semantic relationships. 

As illustrated in Figure \ref{fig:pipeline}, \textit{$M^{THHF}$} consists of two branches with similar structure, modality-specific branch and modality-full branch. The former aims to fuse $\tilde{\textbf{\textit{i}}}_m$ into modality-specific interaction representation $\mathcal{I}$ and the latter aims to fuse $\tilde{\textbf{\textit{h}}}_m$ into the modality-full one $\mathcal{M}$. For each branch, to extract the text-dominated fused features, the textual features are first combined with visual and acoustic features through Text-dominated High-order Fusion blocks, respectively. Then, through the Flat High-order Fusion block, we merge these two text-dominated fused features into a three-modal interaction representation.


After that, we obtain three types of interaction representations $\mathcal{M}$, $\mathcal{S}$, and $\mathcal{I}$. The underlying idea is that $\mathcal{M}$, $\mathcal{S}$, and $\mathcal{I}$ all have three-modal information but encode multimodal interactions in different ways and complement each other. Such an multi-view encoding is able to model sophisticated and informative interactions. Therefore, $\mathcal{S}$, $\mathcal{I}$ and $\mathcal{M}$ are concatenated together, denoted as comprehensive interaction representation $\mathcal{F}_0$. More details can be seen below. For simplicity, we take the modality-specific branch as the example.

\subsubsection{Text-dominated High-order Fusion.} 
To fully model complex interaction among three-modal modality-full and modality-specific representations ($\tilde{\textbf{\textit{h}}}_m$ and $\tilde{\textbf{\textit{i}}}_m$), the outer product is utilized to integrate them. The outer product is able to learn their high-order correlations \cite{he2018outer} and tightly entangles the learning process of multimodal representation elements with a multiplicative interaction between all elements. Therefore, it avoids modality-bias overfitting and has been proved beneficial to the fusion of multimodal features \cite{fukui2016multimodal,lin2015bilinear}. 



\begin{figure}[!t]
\centering
\includegraphics[width=0.48 \textwidth]{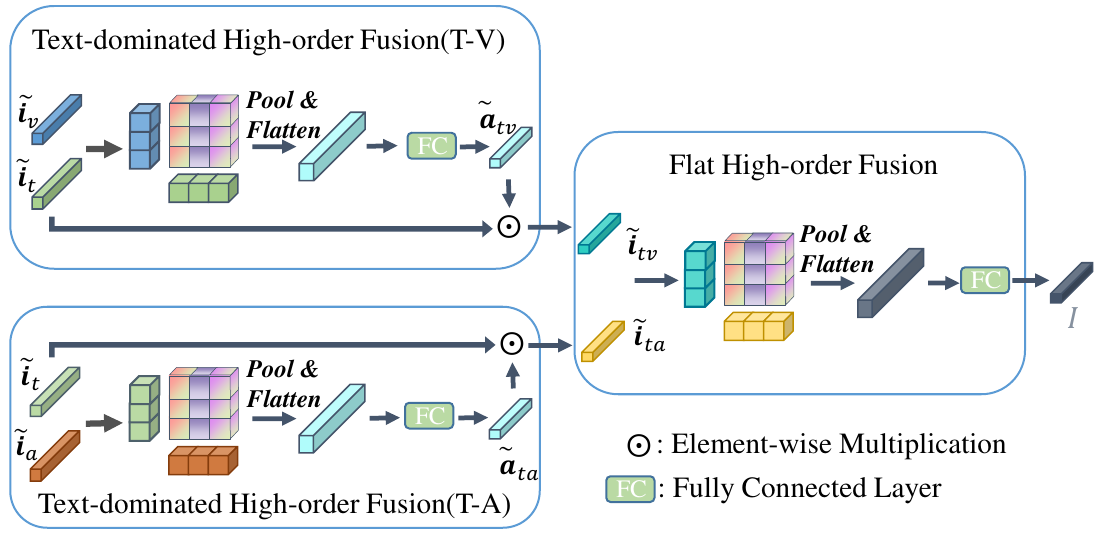}
\caption{Illustration of the modality-specific branch of $M^{THHF}$. Modality-full branch is similar to this.}
\label{fig:TFFH}
\end{figure}

Specifically, $\tilde{\textbf{\textit{i}}}_t$ and $\tilde{\textbf{\textit{i}}}_v$ are first merged into a high-order tensor (16$\times$16) through the outer product. Although outer product tightly entangles the learning processing of multimodal representations, it also brings redundant information. To alleviate this problem, a Max Pooling Layer is introduced to reduce the dimension of high-order tensor into 8$\times$8. Later, the tensor is flattened and reduced into a 16-dimensional attention vector $\tilde{\textbf{\textit{a}}}_{tv}$ which contains the information about the interaction between text and visual modalities. Since empirical research shows that text modality plays the most significant role in mutlimodal emotion analysis, we take text as the main modality and integrate the attention vector $\tilde{\textbf{\textit{a}}}_{tv}$ through the dot product to get the fused representation $\tilde{\textbf{\textit{i}}}_{tv}$, which is formulated as follows:


\begin{equation}
  \mathcal{\tilde{\textbf{\textit{i}}}}_{tv} = \tilde{\textbf{\textit{i}}}_t \odot \textit{FC} (\textit{Flatten}(\textit{Pool}(\tilde{\textbf{\textit{i}}}_t \otimes \tilde{\textbf{\textit{i}}}_v))) \\
\end{equation}
\noindent where $\otimes$ refers to outer product and $\odot$ refers to dot product. Similarly, $\tilde{\textbf{\textit{i}}}_t$ and $\tilde{\textbf{\textit{i}}}_a$ are merged into $\tilde{\textbf{\textit{i}}}_{ta}$. The fusion process is formulated as follows:

\begin{equation}
  \mathcal{\tilde{\textbf{\textit{i}}}}_{ta} = \tilde{\textbf{\textit{i}}}_t \odot \textit{FC} (\textit{Flatten}(\textit{Pool}(\tilde{\textbf{\textit{i}}}_t \otimes \tilde{\textbf{\textit{i}}}_a))) \\
\end{equation}





\begin{table*}[!t]
\centering
\setlength{\tabcolsep}{4.5pt}
\footnotesize
\caption{Performances of emotion analysis on MOSEI and MOSI with InterMulti (Ours) and other baselines. Best results are highlighted in bold. $\bigtriangleup$ means results are excerpted from previous papers; $\dagger$ indicates results are reproduced from open-source code with hyper-parameters provided in original papers. In the column of Acc-2 and F1, The */* refers to two results on the metric of \emph{negative/non-negative} or \emph{negative/positive}, respectively. And $'-'$ denotes that the corresponding terms are not reported in the original works. $C$ refers to the classical features provided by datasets; $B^{T}$, $B^{S}$ and $B^{TS}$ refer to using Text-BERT features, Speech-BERT and both Text-BERT and Speech-BERT features, respectively, with classical features for other modalities. $w/o$ $V$ refers to moving out the visual modality. }
\label{table:moseimosi}
\begin{tabular}{cc|ccccc|cccccc}
\Xhline{0.8pt}

\multicolumn{2}{c|}{\multirow{2}{*}{Method}} & \multicolumn{5}{c|}{MOSEI} & \multicolumn{5}{c}{MOSI} \\ \cline{3-12} 
\multicolumn{2}{c|}{} & \multicolumn{1}{c}{MAE $\downarrow$} & \multicolumn{1}{c}{Corr $\uparrow$} & \multicolumn{1}{c}{Acc-2 $\uparrow$} & \multicolumn{1}{c}{F1 $\uparrow$} & \multicolumn{1}{c|}{Acc-7 $\uparrow$} & \multicolumn{1}{c}{MAE $\downarrow$} & \multicolumn{1}{c}{Corr $\uparrow$} & \multicolumn{1}{c}{Acc-2 $\uparrow$} & \multicolumn{1}{c}{F1 $\uparrow$} & \multicolumn{1}{c}{Acc-7 $\uparrow$} \\
\hline
& MFN(C)$^\bigtriangleup$              & -     & -      & 76.0/-    & 76.0/-   & -     & 0.965 & 0.632  & 77.4/-    & 77.3/-    & 34.1 \\
& MV-LSTM(C)$^\bigtriangleup$    & -     & -      & 76.4/-    & 76.4/-   & -     & 1.019 & 0.601  & 73.9/-    & 74.0/-    & 33.2 \\
& SPT(C)$^\bigtriangleup$ & - & - & -/82.6 & -/82.8 & - & - & - & -/82.8 & -/\textbf{82.9} & - \\
& MulT(C)$^\bigtriangleup$  & 0.580 & 0.703 & -/82.5 & -/82.3 & 51.8 & 0.871 & 0.698 & -/83.0 & -/82.8 & 40.0 \\
& RAVEN(C)$^\bigtriangleup$           & 0.614 & 0.662  & 79.1/-    & 79.5/-   & 50.0  & 0.915 & 0.691  & 78.0/-    & 76.6/-    & 33.2 \\
& MTAG(C)$^\dagger$      & 0.572 & 0.714  & 79.6/82.1 & 79.6/82.4  & 50.9  & 0.852 & 0.715 & 80.6/82.4 & 80.5/82.4 & 38.7 \\
& \textbf{InterMulti}(C)     & \textbf{0.543} &\textbf{ 0.764}  & \textbf{83.4/85.7} &\textbf{83.6/85.8} & \textbf{51.9}  & \textbf{0.793} & \textbf{0.756}  & \textbf{81.1/83.0} &\textbf{80.8}/82.0  & \textbf{42.1} \\    \hline



& LMF($B^{T}$)$^\bigtriangleup$           & 0.623 & 0.677  & -/82.0    & -/82.1   & 48.0  & 0.917 & 0.695  & -/82.5    & -/82.4    & 33.2 \\
& TFN($B^{T}$)$^\bigtriangleup$           & 0.593 & 0.700  & -/82.5    & -/82.1   & 50.2  & 0.901 & 0.698  & -/80.8    & -/80.7    & 34.9 \\ 
& MFM($B^{T}$)$^\bigtriangleup$            & 0.568 & 0.717  & -/84.4    & -/84.3   & 51.3  & 0.877 & 0.706  & -/81.7    & -/81.6    & 35.4 \\ 
& ICCN($B^{T}$)$^\bigtriangleup$          & 0.565 & 0.713   & -/84.2    & -/84.2   & 51.6  & 0.860 & 0.710  & -/83.0    & -/83.0    & 39.0 \\ 
& MISA($B^{T}$)$^\bigtriangleup$          & 0.555 & 0.756  & 83.6/85.5   &83.8/85.3  & 52.2  & 0.783 & 0.761 & 81.8/83.4 & 81.7/83.6 & 42.3 \\
& BBFN($B^{T}$)$^\bigtriangleup$          & 0.557 & 0.767  & -/86.2   & -/\textbf{86.1}  & \textbf{54.8}  & 0.776 & 0.755 & -/84.3 & -/84.3 & 45.0 \\
& MMIM($B^{T}$)$^\dagger$            & 0.530 & 0.766  & 82.4/85.8   & 82.5/85.9  & 54.0  & 0.720 & 0.787 & \textbf{82.8}/84.6 & 82.7/84.6 & 44.8 \\
& MAG-BERT($B^{T}$)$^\dagger$           & 0.543 & 0.755  & 82.5/84.8   & 83.0/84.7  & 52.7  & 0.776 & 0.781 & 82.4/84.4 & 82.5/84.6 & 43.6 \\
& Self-MM($B^{T}$)$^\dagger$         & 0.532 & 0.764  & 82.5/84.9 & 82.5/85.1  & 53.1  & 0.725 & 0.789 & \textbf{82.8}/84.2 & \textbf{82.8}/84.2 & 44.8 \\
& \textbf{InterMulti}($B^{T}$)  &\textbf{0.528} &\textbf{0.776}   & \textbf{85.1}/\textbf{86.3} &\textbf{85.1}/\textbf{86.1} & 53.9 &\textbf{0.714} &\textbf{0.795} &82.1/\textbf{84.8}  &81.9/\textbf{84.8} & \textbf{45.3} \\ \hline
& SSL($B^{TS}$, $w/o$ $V$)$^\bigtriangleup$ & 0.491 & -    & -/88.0    & -/\textbf{88.1}    & 56.0 &\textbf{0.577}  & -     & -/88.3    & -/88.6   & 47.2\\ 
& \textbf{InterMulti} ($B^{TS}$, $w/o$ $V$)  & \textbf{0.483} & \textbf{0.773} & \textbf{87.0/88.3} & \textbf{86.9/88.1} & \textbf{56.5} & 0.586 & \textbf{0.871} & \textbf{87.8/89.2} & \textbf{87.8/89.1} & \textbf{51.5} \\ 
\Xhline{0.8pt}
\end{tabular}
\end{table*}

\subsubsection{Flat High-order Fusion.} 
Next, to further capture the interactions across multiple modalities, we merge two Text-dominated fusion representations $\tilde{\textbf{\textit{i}}}_{tv}$ and $\tilde{\textbf{\textit{i}}}_{ta}$ into a modality-specific interaction representation $\mathcal{I}$ through Flat High-order Fusion. To be specific, we use an outer product following a pooling layer and a fully connected layer to perform fusion, which is formulated as follows:

\begin{equation}
  \mathcal{I} = \textit{FC}(\textit{Flatten} (\textit{Pool}(\tilde{\textbf{\textit{i}}}_{ta} \otimes \tilde{\textbf{\textit{i}}}_{tv}))) \\
\end{equation}
\noindent Similarly, in the modality-full branch, we merged $\tilde{\textbf{\textit{h}}}_{m}$ into a modality-full interaction representation $\mathcal{M}$.

\subsubsection{Comprehensive Fusion.}
Lastly, we take three intermediate multimodal interaction representations, $\mathcal{S}$, $\mathcal{I}$ and $\mathcal{M}$, as inputs to characterize multi-view multimodal interactions from different views:

\begin{enumerate}
\item A modality-shared interaction representation $\mathcal{S}$ modeling three-modal common information; 
\item A three-modal modality-specific interaction representation $\mathcal{I}$ modeling the interaction among three modality-specific representations; 
\item Another three-modal modality-full interaction representation $\mathcal{M}$ integrates three full unimodal representations  $\textbf{\textit{h}}_m$ and models three-modal intact information interaction, while $\mathcal{S}$ and $\mathcal{I}$ model interactions relying on partial information from $\textbf{\textit{h}}_m$. 
\end{enumerate}

The well-designed end-to-end training allows $\mathcal{S, M, I}$ to model distinguishable multimodal interactions according to their specific essences mentioned above. To further unleash the representation capacity of the multimodal interaction for emotion analysis, three intermediate interaction representations $\mathcal{S}$, $\mathcal{I}$ and $\mathcal{M}$ are then concatenated into a comprehensive interaction representation $\mathcal{F}_0$.

\subsection{Training Loss}
Our experiments perform emotion analysis, including emotion classification or sentiment intensity prediction, depending on the labels provided by the datasets (details in Section \ref{sec:experiment}). Specifically, when performing emotion classification, the standard cross-entropy loss is employed as follows:
\begin{equation}
\mathcal{L}_{\text {CE}}=-\frac{1}{N} \sum_{i=0}^{N} \mathrm{y}_{i} \cdot \log \hat{\mathrm{y}}_{i},
\end{equation}

When performing intensity prediction, we take the mean squared error (MSE) loss as follows:
\begin{equation}
\mathcal{L}_{\text {MSE }}=-\frac{1}{N} \sum_{i=0}^{N}\left\|\mathrm{y}_{i}-\hat{\mathrm{y}}_{i}\right\|^{2},
\end{equation}
where $N$ is the mini-batch size and ${\mathrm{y}}_{i}$ and $\hat{\mathrm{y}}_{i}$ are the ground-truth and predicted intensity for $i$-$th$ utterance, respectively.

\section{Experiment}

\begin{table*}[!t]
\footnotesize
\centering
\caption{Comparative results of emotion analysis on IEMOCAP. Best results are highlighted in bold. $C$ refers to the classical features provided by datasets; $B^{TS}$ refers to using both Text-BERT and Speech-BERT features. $w/o$ $V$ refers to moving out the visual modality. 
}
\label{tabel:iemocap}
\begin{tabular}{cc|cccccccc}
\Xhline{0.8pt}
\multicolumn{2}{c|}{Models} & \multicolumn{8}{c}{IEMOCAP} \\ \hline
\multicolumn{2}{c|}{\multirow{2}{*}{Emotions}} & \multicolumn{2}{c}{Happy} & \multicolumn{2}{c}{Angry} & \multicolumn{2}{c}{Sad} & \multicolumn{2}{c}{Neutral} \\ \cline{3-10} 
\multicolumn{2}{c|}{} & Acc-2$\uparrow$ & F1$\uparrow$ & Acc-2$\uparrow$ & F1$\uparrow$ & Acc-2$\uparrow$ & F1$\uparrow$ & Acc-2$\uparrow$ & F1$\uparrow$ \\ \hline
\multirow{8}{*}{\begin{tabular}[c]{@{}c@{}} \end{tabular}} & MV-LSTM(C) & 85.9 & 81.3 & 85.1 & 84.3 & 80.4 & 74.0 & 67.0 & 66.7 \\
 & MARN(C)   & 86.7 & 83.6 & 84.6 & 84.2 & 82.0 & 81.2 & 66.8 & 65.9 \\ 
 & MFN(C) & 86.5 & 84.0 & 85.0 & 83.7 & 83.5 & 82.1 & 69.6 & 69.2 \\           
 & RMFN(C) & 87.5 & 85.8 & 84.6 & 84.2 & 82.9 & 85.1 & 69.5 & 69.1 \\ 
 & RAVEN(C) & 87.3 & 85.8 & 87.3 & 86.7 & 83.4 & 83.1 & 69.7 & 69.3 \\ 
 & TFN(C) & 86.7 & 84.0 & 87.1 & 87.0 & 85.6 & 85.8 & 68.9 & 68.0 \\  
 & LMF(C) & 86.1 & 83.9 & 86.2 & 86.4 & 84.3 & 84.4 & 69.6 & 68.8 \\ 
 & MulT(C) & \textbf{90.7} & \textbf{88.6} & 87.4 & 87.0 & 86.7 & 86.0 & 72.4 & 70.7 \\
 & MFM(C) & 86.7 & 84.7 & 87.0 & 86.7 & 85.7 & 85.7 & 70.3 & 69.9 \\ 
 & ICCN(C) & 87.4 & 84.7 & 88.6 & 88.0 & 86.3 & 85.9 & 69.7 & 68.5 \\  
 & MTAG(C) & - & 86.0 & - & 76.7 & - & 79.9 & - & 64.1 \\  

\begin{tabular}[c]{@{}c@{}}\end{tabular} &\textbf{InterMulti}(C) & 87.4 & 86.3 & \textbf{90.1} & \textbf{89.8} & \textbf{86.8} &\textbf{86.5} &\textbf{74.1} &\textbf{73.9} \\ \hline 

& SSL($B^{TS}$, $w/o$ $V$)  & 84.1 & 83.8 & 93.6 & 93.6 & 90.9 & 90.7 & 81.6 & 81.1 \\ 
& \textbf{InterMulti} ($B^{TS}$, $w/o$ $V$) & \textbf{85.1} &  \textbf{84.3} & \textbf{94.0} &  \textbf{93.9} & \textbf{91.5} &  \textbf{91.3} & \textbf{82.6} &  \textbf{81.2} \\

\Xhline{0.8pt}
\end{tabular}
\end{table*}

Extensive experiments are conducted to evaluate the proposed InterMulti. First, Our InterMulti is validated by comparing to baselines and analyzing on its performance. Secondly, ablation studies are conducted to look into the performance of the InterMulti. 

\textbf{Implementation Details.} The optimization uses the Adam optimizer \cite{kingma2017adam} with learning rate 0.0001. The training duration of each model is governed by an early-stopping strategy with the patience of 10 epochs. The size of the mini-batch is 64. The training relies on one GTX3090 GPU. The partition of the data for training, validation, and testing follows the official setting adopted by most of the previous works. 

\textbf{Datasets.}
Our experiments perform on the widely-used MOSI \citep{zadeh2016multimodal}, MOSEI \citep{zadeh2018multimodal} and IEMOCAP \citep{busso2008iemocap}. MOSI contains 2,199 utterance samples from YouTube, spanning 89 different speakers. Each sample is manually annotated with seven emotional intensities ranging from $-3$ to $+3$, where -3/+3 indicates a strong negative/positive emotion. MOSEI is an extended version of MOSI with unified annotation labels. It includes 23,453 annotated utterance samples, 1,000 distinct speakers, and 250 different topics. The IEMOCAP contains 9,958 utterance samples from professional actors. In IEMOCAP, each utterance sample is labelled with one emotion label, including seven kinds of emotion. Due to the imbalance of some emotion labels, the four categories of Happy, Sad, Anger, and Neutral are most commonly used in the previous works, which is followed by our work. 

Due to the difference in the set of labels in the above datasets, our experiments predict emotion intensity in MOSI/MOSEI and classify emotion labels in IEMOCAP.

\textbf{Preprocessing.} Traditionally, text modality features are GloVe \cite{pennington2014glove} embeddings; for visual modality, previous works use Facet to extract facial expression features \cite{ekman1980facial}; for acoustic modality, COVAREP \cite{degottex2014covarep} is used to extract the features. We call these three features as the \textit{classical} features denoted as $C$. Recent works \cite{rahman2020integrating, MISA} have demonstrated that BERT \cite{devlin2018bert} can provide better features than GloVe for emotion analysis and boost the performance. Later on, \citet{twobert} further introduces Speech-BERT \cite{speechBert} to extract audio representations. For a fair comparison, we also conduct experiments on BERT features and Speech-BERT features denoted as $B^T$ and $B^S$, respectively. 

Additionally, we perform word-level alignment to obtain aligned data by the prevalent tool kit P2FA \cite{P2FA}, which is regularly employed in previous works.

\begin{table}[!t]
\centering
\footnotesize
\caption{Dependency between different decoupled representations based on MOSEI.}
\begin{tabular}{c|cccccc}
\Xhline{0.8pt}
Methods & ${\textbf{\textit{i}}}_a$ - ${\textbf{\textit{i}}}_t$ & ${\textbf{\textit{i}}}_a$ - ${\textbf{\textit{i}}}_v$ & 
${\textbf{\textit{i}}}_t$ - ${\textbf{\textit{i}}}_v$ & 
${\textbf{\textit{i}}}_a$ - ${\textbf{S}}$&
${\textbf{\textit{i}}}_t$ - ${\textbf{S}}$& 
${\textbf{\textit{i}}}_v$ - ${\textbf{S}}$
\\ 
\Xhline{0.8pt}
InterMulti & 4.3 & 4.2 & 4.4 & 4.3 & 4.4 & 2.8\\ 
MISA  & 28.9 & 28.8 & 30.3 & 28.6 & 30.58 & 30.2\\ 
\Xhline{0.8pt}
\end{tabular}
\label{table:smilarity}
\end{table}

\subsection{Comparisons with the State-of-the-arts}

\label{sec:exp}
The baseline methods include MFN \cite{MFN}, MV-LSTM \cite{MVLSTM}, RAVEN \cite{wordsCShift}, MARN \cite{MARN}, RMFN \cite{Multistage}, SPT \cite{cheng2021multimodal}, MCTN \cite{Foundintranslation}, MulT \cite{MulT}, TFN \cite{zadeh2017tensor}, LMF \cite{LMF}, MFM \cite{tsai2018learning}, ICCN \cite{ICCN}, MISA \cite{MISA}, BBFN \cite{BBFN}, MMIM \cite{MMIM}, Self-MM \cite{Self-MM}, MAG-BERT \cite{MAG}, MTAG \cite{MTAG} and SSL\cite{twobert}. As done in their works, MAE (mean absolute error), Corr (Pearson correlation coefficient), Acc-7 (7-class accuracy), Acc-2 (binary accuracy) and F1 score are taken as the metrics on MOSI/MOSEI, while Acc-2 and F1 score on IEMOCAP. Specially, Acc-2 may refer to \emph{negative/non-negative} or \emph{negative/positive} on MOSI/MOSEI while refers to the latter on IEMOCAP. 






\label{sec:experiment}
\begin{table*}[tbp]
\footnotesize
\centering
\caption{Results of the ablation study. All the experiments employ the classical features (C) provided by datasets.} 
\begin{tabular}{cc|ccc|ccc|clclclcl}
\Xhline{0.8pt}
\multicolumn{2}{c|}{Dataset} & \multicolumn{3}{c|}{MOSEI} & \multicolumn{3}{c|}{MOSI} & \multicolumn{8}{c}{IEMOCAP}  \\ 
\hline
\multicolumn{2}{c|}{\multirow{2}{*}{Ablation Study}} & \multirow{2}{*}{MAE $\downarrow$} & \multirow{2}{*}{Corr $\uparrow$} & \multirow{2}{*}{F1 $\uparrow$} & \multirow{2}{*}{MAE $\downarrow$} & \multirow{2}{*}{Corr $\uparrow$} & \multirow{2}{*}{F1 $\uparrow$} &\multicolumn{8}{c}{F1 $\uparrow$}  \\ \cline{9-16} 
\multicolumn{2}{c|}{} &  &  &  &  &  &  & \multicolumn{2}{c}{Happy} & \multicolumn{2}{c}{Angry} & \multicolumn{2}{c}{Sad} & \multicolumn{2}{c}{Neutral}  \\\hline
\textbf{A0} & \textbf{InterMulti}    & \textbf{0.543} & \textbf{0.764} & \textbf{85.8} & \textbf{0.793} & \textbf{0.756} & \textbf{82.0}  & \multicolumn{2}{c}{\textbf{86.3}} & \multicolumn{2}{c}{\textbf{89.8}} & \multicolumn{2}{c}{\textbf{86.5}} & \multicolumn{2}{c}{\textbf{73.9}} \\ \hline
A1 & $\mathcal{S}$ + $\mathcal{M}$             & 0.549 & 0.759 & 83.2 & {0.807} & 0.740 & 81.2 & \multicolumn{2}{c}{85.6} & \multicolumn{2}{c}{88.7} & \multicolumn{2}{c}{82.6} & \multicolumn{2}{c}{71.7} \\
A2 & $\mathcal{S}$ + $\mathcal{I}$               & 0.546 & 0.760 & 85.4 & 0.819 & 0.735 & 79.3 & \multicolumn{2}{c}{81.3} & \multicolumn{2}{c}{88.4} & \multicolumn{2}{c}{83.8} & \multicolumn{2}{c}{69.7}\\
A3 & $\mathcal{M}$   + $\mathcal{I}$               & 0.547 & 0.760 & 85.4 & 0.830 & 0.726 & 81.3  & \multicolumn{2}{c}{85.6} & \multicolumn{2}{c}{{89.0}} & \multicolumn{2}{c}{80.3} & \multicolumn{2}{c}{72.3} \\
A4 & $\mathcal{S}$ only            & 0.546 & 0.759 & 84.8 & 0.841 & 0.717 & 79.6  & \multicolumn{2}{c}{85.1} & \multicolumn{2}{c}{88.8} & \multicolumn{2}{c}{80.1} & \multicolumn{2}{c}{70.0}\\
A5 & $\mathcal{M}$   only            & 0.572 & 0.740 & 85.0 & 0.831 & 0.739 & 81.3  & \multicolumn{2}{c}{85.7} & \multicolumn{2}{c}{88.5} & \multicolumn{2}{c}{83.5} & \multicolumn{2}{c}{70.5} \\
A6 & $\mathcal{I}$   only            & 0.545 & 0.758 & 85.0 & 0.811 & 0.741 & 80.9  & \multicolumn{2}{c}{85.2} & \multicolumn{2}{c}{88.0} & \multicolumn{2}{c}{82.8} & \multicolumn{2}{c}{72.4} \\ \hline
A7 & w/o text          & 0.823 & 0.273 & 67.7 & 1.455 & 0.015 & 59.4 & \multicolumn{2}{c}{83.0} & \multicolumn{2}{c}{86.8} & \multicolumn{2}{c}{78.7} & \multicolumn{2}{c}{65.1}\\ 
A8 & w/o visual        & 0.551 & 0.755 & {85.5} & 0.880 & 0.708 & 80.6& \multicolumn{2}{c}{81.7} & \multicolumn{2}{c}{88.2} & \multicolumn{2}{c}{81.4} & \multicolumn{2}{c}{69.7} \\ 
A9 & w/o acoustic      & 0.553 & 0.752 & 85.0 & 0.873 & 0.719 & 81.7& \multicolumn{2}{c}{84.4} & \multicolumn{2}{c}{85.2} & \multicolumn{2}{c}{81.3} & \multicolumn{2}{c}{69.5} \\ 


\hline 
A10 & ortho. constraint & 0.557 & 0.757 & 85.5 & 0.813 & {0.744} & {80.7} & \multicolumn{2}{c}{85.0} & \multicolumn{2}{c}{87.0} & \multicolumn{2}{c}{82.1} & \multicolumn{2}{c}{71.0}  \\ \hline
A11 & w/o \textit{$M^{THHF}$} & 0.556 & 0.743 & 84.9 & 0.814 & 0.732 & 80.4 & \multicolumn{2}{c}{85.2} & \multicolumn{2}{c}{88.8} & \multicolumn{2}{c}{84.9} & \multicolumn{2}{c}{72.1} \\ 
A12 & w/o outer product & 0.541 & 0.755 & 85.4 & 0.809 & 0.736 & 81.5 & \multicolumn{2}{c}{{85.9}} & \multicolumn{2}{c}{88.6} & \multicolumn{2}{c}{{85.8}} & \multicolumn{2}{c}{{73.6}} \\ 
A13 & w/o text-dominated fusion & 0.551 & 0.759 & 85.1 & 0.806 & 0.731 & 80.1 & \multicolumn{2}{c}{{85.0}} & \multicolumn{2}{c}{88.2} & \multicolumn{2}{c}{{86.1}} & \multicolumn{2}{c}{{71.9}} \\ 
A14 & co-attention & 0.562 & 0.751 & 84.8 & 0.826 & 0.742 & 80.2 & \multicolumn{2}{c}{84.9} & \multicolumn{2}{c}{88.5} & \multicolumn{2}{c}{84.2} & \multicolumn{2}{c}{71.5} \\
A15 & visual-dominated fusion & 0.549 & 0.741 & 85.4 & 0.804 & 0.743 & 81.4 & \multicolumn{2}{c}{{85.5}} & \multicolumn{2}{c}{88.7} & \multicolumn{2}{c}{{85.9}} & \multicolumn{2}{c}{{73.6}} \\ 
A16& acoustic-dominated fusion & 0.569 & 0.752 & 84.6 & 0.812 & 0.731 & 81.3 & \multicolumn{2}{c}{{85.8}} & \multicolumn{2}{c}{89.5} & \multicolumn{2}{c}{{86.1}} & \multicolumn{2}{c}{{73.4}} \\ 
\Xhline{0.8pt}

\end{tabular}
\label{table:Abla}
\end{table*}

\subsubsection{Results.}
To make fair comparisons, our results rely on the identical features with the baselines. As shown in Table \ref{table:moseimosi}, compared with the methods followed ($C$) using the classical features, our results outperform all the baselines on MOSI and MOSEI and almost baselines on IEMOCAP except MulT.
However, our InterMulti outperforms MulT on all the metrics on MOSI and MOSEI and three out of four emotions on IEMOCAP. Looking into those methods using the classical audio and vision features but text-BERT features($B^{T}$), our InterMulti outperforms the baselines on almost all the metrics except Acc-2(\textit{negative/non-negative}) on MOSI and ACC-7 on MOSEI. Particularly, compared to SSL using two-modal features of text-BERT and audio-Bert, our InterMulti ($B^{TS}$) performs better on nine metrics, except for MAE on the smaller dataset of MOSI. The above observations validate our proposed framework InterMulti.  

\begin{figure}[!t]
\centering
\includegraphics[width=0.35 \textwidth]{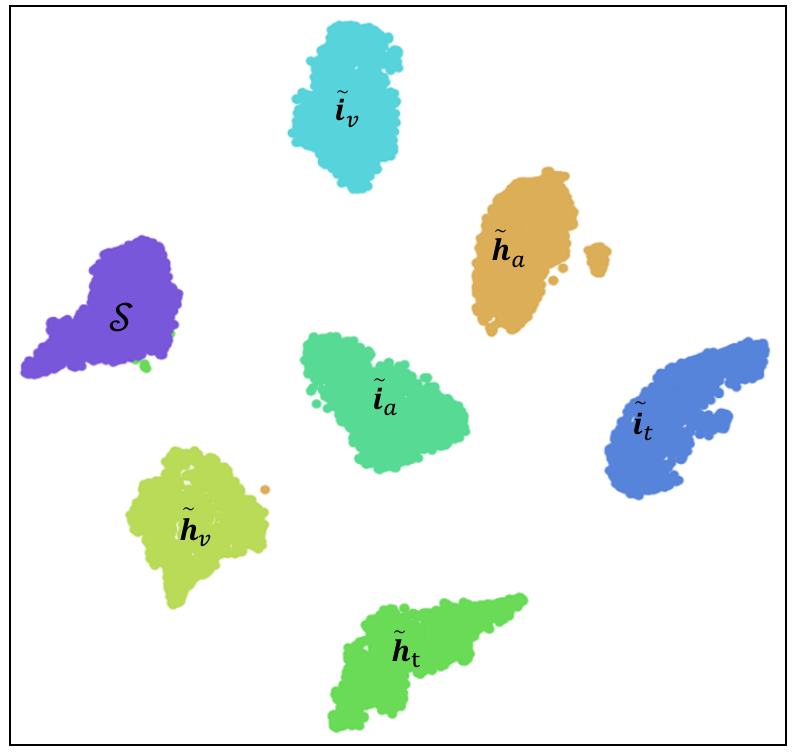}
\caption{The visualization of the representations ${\textbf{\textit{i}}}_t$, ${\textbf{\textit{i}}}_v$, ${\textbf{\textit{i}}}_a$, ${\textbf{\textit{h}}}_t$, ${\textbf{\textit{h}}}_v$, ${\textbf{\textit{h}}}_a$ and $\mathcal{S}$ , based on the test set of MOSEI. The dimension reduction is conducted with tSNE.}
\label{fig:tsne}
\end{figure}

To further obtain insight into our method, we compare the dependency between our decoupling representations with MISA \cite{MISA}. MISA decouples multimodal features by the distribution-based training that requires additional trainable parameters. Since the dot product represents the length of projection from one vector to the other, we employ it to measure the dependency between two vectors. The results in Table \ref{table:smilarity} indicate that our nonparametric decoupling operation leads to distinguishable representations with lower dependency, facilitating encoding different information and improving representation capacity.
Additionally, Figure \ref{fig:tsne} visualizes the learned representations ${\textbf{\textit{i}}}_t$, ${\textbf{\textit{i}}}_v$, ${\textbf{\textit{i}}}_a$, ${\textbf{\textit{h}}}_t$, ${\textbf{\textit{h}}}_v$, ${\textbf{\textit{h}}}_a$ and $\mathcal{S}$, which are fed into the Hierarchical Fusion Module, by tSNE-based \cite{maaten2008visualizing} dimension reduction. Obvious gaps between their distributions can be seen, which indicating that these representations learn to provide different information and contribute to comprehensive fusion.

\subsection{Ablation Study}
\label{sec:ab}

To further validate the effectiveness of our framework and the proposed components, we conduct ablation studies relying on MOSEI, MOSI and IEMOCAP. For simplicity, we only report on the metric of \emph{negative/positive} on F1. The results are reported in Table \ref{table:Abla}.

To confirm the necessity of three kinds of interaction representations in our proposed framework, we conduct ablation studies A1-A6. A1-A6 are conducted by employing one or two representations of {$\mathcal{S}$, $\mathcal{I}$ and $\mathcal{M}$} while keeping other parts the same as A0. The results validate the effectiveness of our proposed framework. The proposed method (A0) with $\mathcal{S, M, I}$ is better than using one or two of them. It proves that these three interaction representations carry on different interaction information and are capable of modeling the complex multimodal interactions together. All $\mathcal{S, M, I}$ cannot be neglected and cannot replace each other. 

To confirm the capability of fitting to each modality, we conduct A7-A9 in Table \ref{table:Abla} by using only two modal information as inputs and ignoring one modality of text, audio or vision, respectively. It is found that A7-A9 are worse than InterMulti (A0) using all three modalities. It means that InterMulti is capable of extracting meaningful but different information from each modality.

To furthermore confirm our decoupling operation, we conduct A10 which replaces self-decoupling operation in InterMulti (A0) by the orthogonal constraints (including the similarity, difference and reconstruction losses) inspired by MISA \cite{MISA}. The result shows that the performance of A10 goes worse in comparison with A0. This observation furthermore validates our decoupling operation in an end-to-end training.

To validate the effectiveness of \textit{$M^{THHF}$}, in A11, we replaces \textit{$M^{THHF}$} by just concatenating three types of representations, $\tilde{\textbf{\textit{i}}}_m$,
$\tilde{\textbf{\textit{h}}}_m$, and $\mathcal{S}$. The result shows that the performance of A11 goes worse without hierarchical fusion, which means that our hierarchical fusion can effectively model the complex interaction between different modalities. We further replace the outer product with concatenation in A12 and the Text-dominated High-order Fusion with Flat High-order Fusion in A13. 
Besides, we bring in co-attentional transformer layers \cite{lu2019vilbert} to replace the outer product in A14. The performance of A12, A13 and A14 also goes worse compared to A0, which means that the outer product is necessary for tightly entangling the elements in three-modal representations. In A15 and A16, we also try visual/acoustic-dominated fusion, which is worse than text-domonated fusion. The observation in A11 to A16 validates the well-designed THHF Module.

All the above experimental results confirm the effectiveness of our InterMulti from three-modal signals, the use of three diverse interaction representations $\mathcal{S, M, I}$, the nonparametric decoupling operation and the text-dominated hierarchical high-order fusion.

\section{Conclusion}
In this paper, we propose a novel framework named InterMulti for multimodal emotion analysis. InterMulti is dedicated to modeling the complex multimodal interactions across multiple modalities in diverse views. Based on that, a novel nonparametric decoupling operation is developed to learn modality-shared/specific interaction representations. It benefits from ease of implementation and no additional parameters or constraints. Additionally, a powerful Text-dominated Hierarchical High-order Fusion (THHF) is proposed to reasonably integrates the diverse representations for emotion analysis. The experiments validate the effectiveness of our InterMulti.

\bibliography{aaai23}
\end{document}